%% file: conll2018.tex
\documentclass[11pt,a4paper]{article}
\usepackage[hyperref]{conll2018}
\usepackage{times}
\usepackage{latexsym}
\usepackage{amsmath}

\usepackage[]{todonotes}   

\usepackage{url}

\newcommand{\is}{$^\ast$}
\newcommand{\ns}{\phantom{$^\ast$}}

\makeatletter
\def\@xfootnote[#1]{%
  \protected@xdef\@thefnmark{#1}%
  \@footnotemark\@footnotetext}
\makeatother

\aclfinalcopy 

\title{Sentence-Level Fluency Evaluation:\\References Help, But Can Be Spared!}

\author{  
  Katharina Kann\thanks{*This research was carried out while the first author was interning at Google.}\\ 
 Center for Data Science\\New York University\\New York, USA\\
  \texttt{kann@nyu.edu} \\
  \And
  Sascha Rothe\\
  Google Research\\
  Zurich, Switzerland\\
  \texttt{rothe@google.com} \\
  \AND
  Katja Filippova \\
  Google Research\\
  Berlin, Germany\\
  \texttt{katjaf@google.com} \\
}

\date{}

\begin{document}
\maketitle
\begin{abstract}
Motivated by recent findings on the probabilistic modeling of acceptability judgments, 
we propose syntactic log-odds ratio (SLOR), a normalized
language model score, as a metric for referenceless fluency evaluation of natural language generation output at the sentence level.
We further introduce WPSLOR, a novel WordPiece-based version,
which harnesses a more compact language model.
Even though word-overlap metrics like ROUGE are computed with the help of hand-written references,
our referenceless methods obtain a significantly higher
correlation with human fluency scores on a benchmark dataset of compressed sentences.
Finally, we present ROUGE-LM, a reference-based metric which is a natural extension of WPSLOR to the case of available references. We show that ROUGE-LM yields a significantly higher correlation 
with human judgments 
than all baseline metrics, including WPSLOR on its own.
\end{abstract}

\section{Introduction}
Producing sentences which are perceived as natural by a human 
addressee---a property which we will denote as \textit{fluency}\footnote{Alternative names include \textit{naturalness}, 
\textit{grammaticality} or \textit{readability}. 
Note that the exact definitions of all those terms vary slightly throughout the literature.}
throughout this
paper
---is a crucial goal of all natural language generation (NLG) systems:
it makes interactions more natural, avoids misunderstandings and,
overall, leads to higher user satisfaction and user trust \cite{usertrust2018}.
Thus, 
fluency evaluation
is important, e.g., during system development,
or for filtering unacceptable generations at application time.
However, fluency evaluation of NLG systems constitutes a hard challenge:
systems
are often not limited to reusing words from the input, but can generate 
in an \textit{abstractive} way. Hence, it is not guaranteed that a correct output will match any of a finite number of given references.
This results in difficulties for current reference-based evaluation,
especially of fluency,  
causing word-overlap metrics like ROUGE \cite{lin-och:2004:ACL}
to correlate only weakly with human 
judgments \cite{toutanova2016dataset}. 
As a result, fluency evaluation of NLG is often
done manually, which is costly and time-consuming.

\begin{table}[t!]
\centering
\setlength{\tabcolsep}{4.pt} 
 \begin{tabular}{| l c |}\hline
 If access to a synonym dictionary is & \\ 
 likely to be of use, then this package may & 3 \\
 be of service. & \\\vspace{-0.14cm}
 & \\
 Participants are invited to submit a set & \\
 pair do domain name that is already & 1.6 \\ 
 taken along with alternative. & \\\vspace{-0.14cm}
 &\\
 Even \$15 was The HSUS. & 1 \\ \hline
 \end{tabular}
 \caption{Example compressions from our dataset with their fluency scores; scores in $[1, 3]$, higher is better.\label{tab:example}}
\end{table}
Evaluating sentences on their fluency, on the other hand, is a linguistic ability of humans which has been the subject
of a decade-long debate in cognitive science.
In particular, 
the question has been raised whether the grammatical knowledge that underlies
this ability is probabilistic
or categorical in nature \cite{chomsky2002syntactic,manning2003probabilistic,sprouse2007continuous}.  
Within this context,
\newcite{lau2017grammaticality}
have recently shown that neural language models (LMs)
can be used for modeling human ratings of acceptability.
Namely, they found 
SLOR \cite{pauls2012large}---sentence log-probability which is normalized by
unigram log-probability and sentence length---to correlate
well with acceptability judgments at the sentence level.

However, to the best of our knowledge, these insights have so far gone disregarded by the natural language processing (NLP) community.
In this paper, we investigate the practical implications of \newcite{lau2017grammaticality}'s findings for fluency evaluation of NLG, using the task of automatic compression \cite{Knight:2000:SSS:647288.721086,mcdonald2006discriminative} 
as an example (cf. Table \ref{tab:example}).
Specifically, we test our hypothesis that
 SLOR should be a suitable metric for
evaluation of compression fluency which 
(i) does not rely on references; 
(ii) can naturally be applied at the sentence level (in contrast to the system level); and
(iii) does not need human fluency annotations of any kind.
In particular the first aspect, i.e., SLOR not needing references, makes it a promising candidate for automatic evaluation. Getting rid of human references has practical importance in a variety of settings, e.g., if references are unavailable due to a lack of resources for annotation, or if obtaining references is impracticable. The latter would be the case, for instance, when filtering system outputs at application time.

We further introduce WPSLOR, a novel, WordPiece \cite{wu2016google}-based version of SLOR, which drastically reduces model size and training time.
Our experiments show that both approaches correlate better with human judgments
than traditional word-overlap metrics, even though the latter do rely on reference compressions.
Finally,
investigating the case of available references and how to incorporate them, we combine WPSLOR and ROUGE
to ROUGE-LM, a novel reference-based metric,
and increase the correlation with human fluency ratings even further.

\paragraph{Contributions. }
To summarize, we make the following contributions:
\begin{enumerate}
\item We empirically show that SLOR is a good referenceless metric for the evaluation of NLG fluency at the sentence level.
\item We introduce WPSLOR, a WordPiece-based version of SLOR, which disposes of a more compact LM without a significant loss of performance. 
\item We propose ROUGE-LM, a reference-based metric, which achieves a significantly higher correlation with human fluency judgments than all other metrics in our experiments.
\end{enumerate}

\section{On Acceptability}
Acceptability judgments, i.e., speakers' judgments of the well-formedness of sentences, have been the basis of much linguistics research \cite{chomsky1964aspects,schutze2016empirical}: a speakers intuition about a sentence is used to draw conclusions about a language's rules.
Commonly, ``acceptability'' is used synonymously with ``grammaticality'', and speakers are in practice asked for grammaticality judgments or acceptability judgments interchangeably. Strictly speaking, however, a sentence can be unacceptable, even though it is grammatical -- a popular example is Chomsky's phrase ``Colorless green ideas   sleep furiously.'' \cite{chomsky2002syntactic} In turn, acceptable sentences can be ungrammatical, e.g., in an informal context or in poems \cite{newmeyer1983grammatical}.

Scientists---linguists, cognitive scientists, psychologists, and NLP researcher alike---disagree about how to represent 
human linguistic abilities. One subject of debates are
acceptability judgments: while, for many, acceptability is a binary condition on membership in a set of well-formed sentences \cite{chomsky2002syntactic}, others assume that it is gradient in nature \cite{heilman2014predicting,toutanova2016dataset}.
Tackling this research question, \newcite{lau2017grammaticality} 
aimed at modeling human acceptability judgments automatically, with the goal to gain insight into the nature of 
human perception of acceptability. In particular, they tried to answer the question: Do humans judge acceptability on a  gradient scale?
Their experiments showed a strong correlation between human 
judgments and 
normalized sentence log-probabilities under a variety of LMs for artificial data they had created by translating and back-translating
sentences with neural models. While they tried different types of LMs, best results were obtained for neural models, namely recurrent neural networks (RNNs).

In this work, we investigate if approaches which have proven successful for modeling acceptability can 
be applied to the NLP problem of automatic fluency evaluation. 

\section{Method}
In this section, we first describe SLOR and the intuition behind this score. Then, we introduce WordPieces, before explaining how we combine the two.

\subsection{SLOR} 
SLOR assigns to a sentence $S$ a score which consists of its log-probability under a given LM,
normalized by unigram log-probability and length:
\begin{align}
    \text{SLOR}(S) = &\frac{1}{|S|} (\ln(p_M(S)) \\\nonumber
                     &- \ln(p_u(S)))
\end{align}
where $p_M(S)$ is the probability assigned to the sentence under the LM. 
The unigram probability $p_u(S)$ of the sentence is calculated as
\begin{equation}
    p_u(S) = \prod_{t \in S}p(t)
\end{equation}
with $p(t)$ being the unconditional probability of a token $t$, i.e., given no context.

The intuition behind subtracting unigram log-probabilities is that a token which is rare on its own
(in contrast to being rare at a given position in the sentence)
should not bring down the sentence's rating.
The normalization by sentence length is necessary in order to not prefer shorter sentences over equally fluent longer ones.\footnote{Note that the sentence log-probability which is normalized by sentence length corresponds to the negative cross-entropy.}
Consider, for instance, the following pair of sentences:
\begin{align}
\textrm{(i)} ~ ~ &\textrm{He is a citizen of France.}\nonumber\\
\textrm{(ii)} ~ ~ &\textrm{He is a citizen of Tuvalu.}\nonumber
\end{align}
Given that both sentences are of equal length and assuming that France appears more often in a given LM training set than Tuvalu, the length-normalized log-probability of sentence (i) under the LM would most likely be higher than
that of sentence (ii). However, since both sentences are equally fluent, we expect
taking each token's unigram probability into account to lead to a more suitable score for our purposes.

We calculate the probability of a sentence with a long-short term memory (LSTM, \mbox{\newcite{hochreiter1997long}}) LM, i.e., a special type of RNN LM, which has been trained on a large 
corpus. More details on LSTM LMs can be found, e.g., in \newcite{sundermeyer2012lstm}.
The unigram probabilities for SLOR are estimated using the same corpus. 

\subsection{WordPieces}
Sub-word units like WordPieces \cite{wu2016google} are getting increasingly important in NLP.
They 
constitute a compromise between characters and words: On the one hand,
they yield a smaller vocabulary, which reduces model size and
training time, and improve handling of rare words, since those are partitioned into 
more frequent segments.
On the other hand, they contain more information than characters. 

WordPiece models are estimated
using a data-driven approach which maximizes the LM likelihood
of the training corpus as described in \newcite{wu2016google} and \newcite{6289079}.

\begin{table}[t!]
\centering
\setlength{\tabcolsep}{4.5pt} 
 \begin{tabular}{l | c c c c } 
 &\textbf{ILP} & \textbf{NAMAS} & \textbf{SEQ2SEQ} & \textbf{T3} \\\hline
 fluency & 2.22 & 1.30 & 1.51 & 1.40\\
 \end{tabular}
 \caption{\label{tab:system_average} Average fluency ratings for each compression system in the dataset by \newcite{toutanova2016dataset}.}
\end{table}
\subsection{WPSLOR}
We propose a novel version of SLOR, by incorporating a LM which is trained on a corpus which has been split by a
WordPiece\footnote{https://github.com/google/sentencepiece} model. This leads to a smaller vocabulary, resulting in a LM with less parameters, which is faster to train (around 12h compared to roughly 5 days for the word-based version in our experiments).
We will refer to the word-based SLOR as WordSLOR and to our newly proposed WordPiece-based version as WPSLOR.

\section{Experiment}
Now, we present our main experiment, in which we assess the performances of WordSLOR and WPSLOR as fluency evaluation metrics. 

\subsection{Dataset}
We experiment on the compression dataset by \newcite{toutanova2016dataset}.
It contains single sentences and two-sentence paragraphs from 
the Open American National Corpus (OANC),
which belong to 4 genres: \textit{newswire}, \textit{letters}, \textit{journal}, and \textit{non-fiction}.
Gold references are manually created 
and the outputs of 4 
compression systems
(ILP (extractive), NAMAS (abstractive), SEQ2SEQ (extractive), and T3 (abstractive);
cf. \newcite{toutanova2016dataset} for details) for the test data
are provided. Each example has 3 to 5 independent human ratings for content and fluency. 
We are interested in the latter, which is rated 
on an ordinal scale from 1 (disfluent) through 3 (fluent).
We experiment on the $2955$ system outputs for the test split.

Average  
fluency scores per system are shown in Table \ref{tab:system_average}. 
As can be seen, ILP produces the best output. 
In contrast, NAMAS is the worst system for fluency. 
In order to be able to judge the reliability of the human annotations, we follow the procedure suggested by \newcite{TACL732} and used by \newcite{toutanova2016dataset}, and compute the quadratic weighted $\kappa$ \cite{cohen1968weighted} for the human fluency scores of the system-generated compressions as $0.337$.

\subsection{LM Hyperparameters and Training}
We train our LSTM
LMs on the English Gigaword corpus \mbox{\cite{parker2011english}}, which consists of news data.

The hyperparameters of all LMs are tuned using perplexity on a held-out part of Gigaword, since we expect LM perplexity and final evaluation performance of WordSLOR and, respectively, WPSLOR to correlate.
Our best networks consist of two layers with 512 hidden units each, and are trained for $2,000,000$
steps with a minibatch size of 128. For optimization, we employ ADAM \mbox{\cite{kingma2014adam}}.

\subsection{Baseline Metrics}
\paragraph{ROUGE-L. }
Our first baseline is ROUGE-L \cite{lin-och:2004:ACL}, 
since it is the most commonly used metric for compression tasks. ROUGE-L 
measures the similarity of two sentences based on their longest common subsequence.
Generated and reference compressions are tokenized and lowercased.
For multiple references, we only make use of the one with the highest score for each example.

\paragraph{N-gram-overlap metrics. }
We compare to the best n-gram-overlap metrics from \newcite{toutanova2016dataset};
combinations of linguistic units (bi-grams (LR2) and tri-grams (LR3)) and scoring measures (recall (R) and F-score (F)). 
With multiple references, we consider the union of the sets of n-grams. 
Again, generated and reference compressions are tokenized and lowercased.

\paragraph{Negative cross-entropy. } We further compare to the negative LM cross-entropy, i.e., the log-probability which is only normalized by sentence length.
The score of a sentence $S$ is calculated as 
\begin{equation}
\text{NCE}(S) = \tfrac{1}{|S|} \ln(p_M(S))
\end{equation}
with $p_M(S)$ being the probability assigned to the sentence by a LM.
We employ the same LMs as for SLOR, i.e., LMs trained on words (WordNCE)
and WordPieces (WPNCE).

\paragraph{Perplexity.} Our next baseline is perplexity, 
which corresponds to the exponentiated cross-entropy:
\begin{equation}
\text{PPL}(S) = \exp(-\text{NCE}(S))
\end{equation}

\paragraph{About BLEU.} Due to its popularity, we also performed initial experiments with BLEU \cite{papineni-EtAl:2002:ACL}. Its correlation with human scores was so low that we do not consider it  
in our final experiments.

\subsection{Correlation and Evaluation Scores}
\paragraph{Pearson correlation. }
Following earlier work \cite{toutanova2016dataset}, we evaluate our metrics using Pearson correlation with human judgments. 
It is defined as the covariance divided by the product of the standard deviations:
\begin{equation}
    \rho_{X,Y} = \frac{\text{cov}(X,Y)}{\sigma_X \sigma_Y}
\end{equation}

\paragraph{Mean squared error. } 
Pearson
cannot accurately judge a metric's performance for sentences of very similar quality, i.e., in the extreme case of rating outputs of identical quality, the correlation is either not defined or $0$, caused by noise of the evaluation model.
Thus, we additionally evaluate using mean squared error (MSE), which is defined as
the squares of residuals after a linear transformation, divided by the sample size:
\begin{equation}
    \text{MSE}_{X,Y} = \underset{f}{\min}\frac{1}{|X|}\sum\limits_{i = 1}^{|X|}{(f(x_i) - y_i)^2}
\end{equation}
with $f$ being a linear function.
Note that, since MSE is invariant to linear transformations of $X$ but not of $Y$, it is a non-symmetric quasi-metric.
We apply it with $Y$ being the human ratings.
An additional advantage as compared to Pearson is that it has an interpretable meaning: the expected error made by a given metric as compared to the human rating.


\begin{table}[t!]
\centering
\setlength{\tabcolsep}{6.2pt} 
 \begin{tabular}{l | c | c c } 
 \textbf{metric} & \textbf{refs} & \textbf{Pearson} & \textbf{MSE} \\ \hline
 {WordSLOR} & \textbf{0} & \textbf{0.454\ns} & \textbf{0.261}\ns\\
 {WPSLOR} & \textbf{0} & 0.437\ns &  0.267\ns \\ \hline
 WordNCE & \textbf{0} & 0.403\is & 0.276\is \\
 WPNCE & \textbf{0} & 0.413\is & 0.273\is \\ \hline
 {WordPPL} & \textbf{0} & 0.325\is & 0.295\is \\
 {WPPPL} & \textbf{0} & 0.344\is &  0.290\is \\ \hline
 {ROUGE-L-mult} & $3-5$& 0.429\is & 0.269\ns \\
 {LR3-F-mult} & $3-5$ & 0.405\is & 0.275\is\\
 {LR2-F-mult} & $3-5$ & 0.375\is & 0.283\is \\
 {LR3-R-mult} & $3-5$ & 0.412\is & 0.273\is \\\hline
 {ROUGE-L-single} & 1 & 0.406\is & 0.275\is \\\hline
 \end{tabular}
 \caption{\label{tab:overallresults} Pearson correlation (higher is better) and MSE (lower is better) for all metrics; best results in bold; \textit{refs}=number of references used to compute the metric.}
\end{table}
\begin{table*}[h!]
\centering
\setlength{\tabcolsep}{4.5pt} 
 \begin{tabular}{l | c | c c c c | c c c c }
 & & \multicolumn{4}{|c|}{Pearson} & \multicolumn{4}{|c}{MSE} \\
 & \textbf{refs} & {}{\textbf{ILP}} & {}{\textbf{NAMAS}} & {}{\textbf{S2S}} & {}{\textbf{T3}} & {}{\textbf{ILP}} & {}{\textbf{NAMAS}} & {}{\textbf{S2S}} & {}{\textbf{T3}} 
 \\
 \hline\hline
  \# samples & & {}{679} & {}{762} & {}{767} & {}{747} & {}{679} & {}{762} & {}{767} & {}{747} 
 \\
 \hline\hline
 WordSLOR & \textbf{0}& 0.363\is& 0.340\is& \textbf{0.257\ns}& 0.343\ns& 0.307\is& 0.104\ns& \textbf{0.161\ns}& 0.174\ns\\
 WPSLOR & \textbf{0}& 0.417\is& 0.312\is& 0.201\is& \textbf{0.360\ns}& 0.292\is& 0.106\is& 0.166\ns& \textbf{0.172\ns} \\
 \hline
 WordNCE &\textbf{0} & 0.311\is& 0.270\is& 0.128\is& 0.342\ns & 0.319\is& 0.109\is& 0.170\is& 0.174\ns \\
 WPNCE &\textbf{0} & 0.302\is& 0.258\is& 0.124\is& 0.357\ns& 0.322\is& 0.110\is& 0.170\is& \textbf{0.172\ns} \\
 \hline
 ROUGE-L-mult & $3-5$& 0.471\ns& \textbf{0.392\ns}& 0.013\is& 0.256\is& 0.275\ns& \textbf{0.100\ns}& 0.173\is& 0.184\is \\
 LR3-F-mult & $3-5$& \textbf{0.489\ns}& 0.266\is& 0.007\is& 0.234\is& \textbf{0.269\ns}& 0.109\is& 0.173\is& 0.187\is \\
 LR2-F-mult & $3-5$& 0.484\ns& 0.213\is& -0.013\is& 0.236\is& 0.271\ns& 0.112\is& 0.173\is& 0.186\is \\
 LR3-R-mult & $3-5$& 0.473\ns& 0.246\is& -0.002\is& 0.232\is& 0.275\is& 0.111\is& 0.173\is& 0.187\is\\
 \hline
 ROUGE-L-single & 1 & 0.363\is& 0.308\is& 0.008\is& 0.263\is& 0.307\is& 0.107\is& 0.173\is& 0.184\is \\
 \hline
 \end{tabular}
 \caption{\label{tab:analysis}  Pearson correlation (higher is better) and MSE (lower is better), reported by compression system; best results in bold;
 \textit{refs}=number of references used to compute the metric.}
\end{table*}
\subsection{Results and Discussion}
As shown in Table \ref{tab:overallresults},
WordSLOR and WPSLOR correlate best with human judgments:
WordSLOR (respectively WPSLOR) has a $0.025$ (respectively $0.008$)
higher Pearson correlation
than the best word-overlap metric ROUGE-L-mult,
even though the latter requires multiple reference compressions.
Furthermore, if we consider with ROUGE-L-single a setting with a single given reference,
the distance to WordSLOR increases to $0.048$ 
for Pearson correlation. Note that, since having a single reference is very common, this result is highly relevant for practical applications.
Considering MSE, the top two metrics are still WordSLOR and WPSLOR, with a $0.008$ and, respectively,
$0.002$ lower error than the third best metric, ROUGE-L-mult.
\footnote[]{*Significantly worse than best (bold) result with $p < 0.05$; one-tailed; Fisher-Z-transformation for Pearson, two sample t-test for MSE.}

Comparing WordSLOR and WPSLOR, we find no significant differences: 
$0.017$ 
for Pearson and $0.006$ for MSE.
However, WPSLOR uses a more compact LM and, hence, 
has a shorter training time,
since the vocabulary is smaller
($16,000$ vs. $128,000$ tokens). 

Next, we find that
WordNCE and WPNCE perform roughly on par with 
word-overlap metrics. This is interesting, since they, in contrast to traditional metrics, do not require reference compressions. 
However, their correlation with human fluency judgments is strictly lower than that of their respective SLOR counterparts. 
The difference between WordSLOR and WordNCE is bigger than that between WPSLOR and
WPNCE.
This might be due to accounting for differences in frequencies being 
more important for words than for WordPieces.
Both WordPPL and WPPPL clearly underperform as compared to all other metrics in our experiments.

The traditional word-overlap metrics all perform
similarly. 
ROUGE-L-mult and {LR2-F-mult} are best and worst, respectively.

\subsection{Analysis I: Fluency Evaluation per Compression System}
\label{subsec:ana_system}
The results per compression system (cf. Table \ref{tab:analysis}) look different from the correlations in Table \ref{tab:overallresults}: Pearson 
and MSE are both lower. This is due to the outputs of each given system being of comparable quality.
Therefore, the datapoints are similar and, thus, easier to fit for the linear function used for MSE.
Pearson, in contrast, is lower due to its invariance to linear transformations of both variables.
Note that this effect is smallest for ILP, which has 
uniformly distributed targets ($\text{Var}(Y) = 0.35$ vs. $\text{Var}(Y) = 0.17$ for SEQ2SEQ).

Comparing the metrics, the two SLOR approaches perform best for SEQ2SEQ and T3. In particular, they outperform the best word-overlap metric baseline by $0.244$ and $0.097$ Pearson correlation as well as $0.012$ and $0.012$ MSE, respectively. Since T3 is an abstractive system, 
we can conclude that WordSLOR and WPSLOR are applicable even for systems that are not limited to make use of a fixed repertoire of words.

For ILP and NAMAS, word-overlap metrics obtain best results. The differences in performance, however, are with a maximum difference of $0.072$ for Pearson and ILP much smaller than for SEQ2SEQ. Thus, while the differences are significant, word-overlap metrics do not outperform our SLOR approaches by a wide margin.
Recall, additionally, that word-overlap metrics rely on references being available, while our proposed approaches do not require this.

\subsection{Analysis II: Fluency Evaluation per Domain}
\label{subsec:ana_domain}
Looking next at the correlations for all models but different domains (cf. Table \ref{tab:analysis_domain}), 
we first observe that
the results across domains are similar, i.e., we do not observe the same effect as in Subsection \ref{subsec:ana_system}. 
This is due to the distributions of scores being uniform ($\text{Var}(Y) \in [0.28, 0.36]$).

Next, we focus on an important question: How much does the performance of our SLOR-based metrics depend on the domain, given that the respective LMs are trained on Gigaword, which consists of news data?

Comparing the evaluation performance for individual metrics, we observe that, except for \textit{letters}, WordSLOR and WPSLOR perform best across all domains: they outperform the best word-overlap metric by at least $0.019$ and at most $0.051$ Pearson correlation, and at least $0.004$ and at most $0.014$ MSE. The biggest difference in correlation is achieved for the \textit{journal} domain.
Thus, clearly even LMs which have been trained on out-of-domain data 
obtain competitive performance for fluency evaluation. 
However, a domain-specific LM might additionally improve the metrics' correlation with human judgments.
We leave a more detailed analysis of the importance of the training data's domain for future work.

\begin{table*}[h!]
\centering
\setlength{\tabcolsep}{4.5pt} 
 \begin{tabular}{l | c | c c c c | c c c c }
 & & \multicolumn{4}{|c|}{Pearson} & \multicolumn{4}{|c}{MSE}  \\
 & \textbf{refs}
 & {}{\textbf{letters}} & {}{\textbf{journal}} & {}{\textbf{news}} & {}{\textbf{non-fi}} & {}{\textbf{letters}} & {}{\textbf{journal}} &
 {}{\textbf{news}} & {}{\textbf{non-fi}}
 \\
 \hline\hline
  \# samples & 
 & {}{640} & {}{999} & {}{344} & {}{972} & {}{640} & {}{999} &
 {}{344} & {}{972}
 \\
 \hline\hline
 WordSLOR & \textbf{0}&  0.452\ns& \textbf{0.453\ns}& \textbf{0.403\ns}& \textbf{0.484\ns}& 0.258\ns& \textbf{0.250\ns}& \textbf{0.234\ns}& \textbf{0.278\ns} \\
 WPSLOR & \textbf{0}& 0.435\is& 0.415\is& 0.389\ns& 0.483\ns& 0.263\ns& 0.260\ns& 0.237\ns& 0.278\ns \\
 \hline
 WordNCE &\textbf{0} & 0.395\is& 0.412\is& 0.342\is& 0.425\is& 0.273\is& 0.261\is& 0.247\ns& 0.297\is \\
 WPNCE &\textbf{0} &  0.424\is& 0.398\is& 0.363\ns& 0.460\ns& 0.266\is& 0.265\is& 0.243\ns& 0.286\ns \\
 \hline
 ROUGE-L-mult & $3-5$& \textbf{0.487\ns}& 0.382\is& 0.384\ns& 0.451\is& \textbf{0.247\ns}& 0.269\is& 0.238\ns& 0.289\ns \\
 LR3-F-mult & $3-5$& 0.404\is& 0.402\is& 0.278\is& 0.439\is& 0.271\is& 0.264\is& 0.258\is& 0.293\ns \\
 LR2-F-mult & $3-5$& 0.390\is& 0.363\is& 0.292\is& 0.395\is& 0.275\is& 0.273\is& 0.256\is& 0.306\is \\
 LR3-R-mult & $3-5$& 0.420\is& 0.395\is& 0.272\is& 0.453\ns& 0.267\is& 0.266\is& 0.259\is& 0.288\ns \\
 \hline
 ROUGE-L-single & 1 & 0.453\ns& 0.347\is& 0.335\is& 0.450\is& 0.258\is& 0.277\is& 0.248\ns& 0.289\ns \\
 \hline
 \end{tabular}
 \caption{\label{tab:analysis_domain}  Pearson correlation (higher is better) and MSE (lower is better), reported by domain of the original sentence or paragraph; best results in bold;
 \textit{refs}=number of references used to compute the metric.}
\end{table*}
\section{Incorporation of Given References}
\begin{table*}[t!]
\centering
\setlength{\tabcolsep}{6.8pt} 
 \begin{tabular}{| c  l |} 
 \hline
 \textbf{model} & \textbf{generated compression} \\ \hline
 ILP & Objectives designed to lead incarcerated youth to an understanding of grief and loss\\
 & related influences on their behavior.\\
 ILP & In Forster's A Passage to India is created.\\
 SEQ2SEQ & Jogged my thoughts back to Muscat Ramble. \\
 SEQ2SEQ & Between Sagres and Lagos, pleasant beach with fishing boats, and a market. \\
 T3 & Your support of the Annual Fund maintaining the core values in GSAS the ethics. \\\hline
 \end{tabular}
 \caption{\label{tab:weird_ratings} Sentences for which raters were unsure if they were perceived as problematic due to fluency or content issues, together with 
 the model which generated them.}
\end{table*}
ROUGE was shown to correlate well with ratings of a generated text's content or meaning at the sentence level \cite{toutanova2016dataset}.  We further expect
content and fluency ratings to be correlated. In fact, sometimes it is difficult to distinguish which one is problematic: to illustrate this, we show some extreme examples---compressions which got the highest fluency rating and
the lowest possible content rating by at least one rater, but the lowest fluency score and the highest content score by
another---in Table \ref{tab:weird_ratings}.
We, thus, hypothesize that ROUGE should contain information about fluency which 
is complementary to SLOR,
and want to make use of references for fluency evaluation, if available.
In this section, we experiment with two \emph{reference-based} metrics -- one trainable one, and one that can be used without fluency annotations, i.e., in the same settings as pure word-overlap metrics.

\subsection{Experimental Setup}
First, we assume a setting in which we have the following available: (i) system outputs whose fluency is to be evaluated, (ii) reference generations for evaluating system outputs, (iii) a small set of system outputs with references, which has been annotated for fluency by human raters, and (iv) a large unlabeled corpus for training a LM.
Note that available fluency annotations are often uncommon in real-world scenarios; the reason we use them is that they allow for a proof of concept.
In this setting, we train scikit's \cite{scikit-learn} support vector regression
model (SVR)
with the default parameters on predicting fluency, given WPSLOR and ROUGE-L-mult.
We use $500$ of our total $2955$ examples for each of training and development,
and the remaining $1955$ for testing. 

Second, we simulate a setting in which we have only access to (i) system outputs which should be evaluated on fluency, (ii) reference compressions, and (iii) large amounts of unlabeled text.
In particular, we assume to not have fluency ratings
for system outputs, which makes training a regression model impossible.
Note that this is the standard setting in which word-overlap metrics are applied.
Under these conditions, we propose to normalize both given scores by mean and variance, and to simply add them up. We call this new reference-based metric ROUGE-LM.
In order to make this second experiment comparable to the SVR-based one, we use the same $1955$ test examples.

\begin{table}[]
\centering
\setlength{\tabcolsep}{.92pt} 
 \begin{tabular}{l | c | c c | c c} 
 &\textbf{metric} & \textbf{refs} & \textbf{train?} & \textbf{Pearson} & \textbf{MSE} \\  
 \hline
  1 &\textbf{SVR}: & $3-5$& yes & \textbf{0.594} & \textbf{0.217} \\
  &ROUGE+WPSLOR & & & & \\
 2 & \textbf{ROUGE-LM} &$3-5$ & no & 0.496 & 0.252 \\ \hline
 3 & ROUGE-L-mult & $3-5$& no & 0.430 &0.273 \\
 4 & WPSLOR & 0 & no & 0.439 & 0.270 \\ \hline
 \end{tabular}
 \caption{\label{results:content}Combinations; all differences except for 3 and 4 are statistically significant; \textit{refs}=number of references used to compute the metric; ROUGE=ROUGE-L-mult; best results in bold.}
\end{table}
\subsection{Results and Discussion}
Results are shown in Table \ref{results:content}.
First, we can see that using SVR (line 1) to combine ROUGE-L-mult and WPSLOR outperforms both individual scores (lines 3-4) by a large margin.
This serves as a proof of concept: the information contained in the two approaches is indeed complementary.

Next, we consider the setting where only references and no annotated examples are available.
In contrast to SVR (line 1), ROUGE-LM (line 2) has only the same requirements as conventional word-overlap metrics (besides a large corpus for training the LM, which is easy to obtain for most languages).
Thus, it can be used in the same settings as other word-overlap metrics.
Since ROUGE-LM---an uninformed combination---performs significantly better than both ROUGE-L-mult and WPSLOR on their own,
it should be the metric of choice for evaluating fluency with given references.

\section{Related Work}
\subsection{Fluency Evaluation}
Fluency evaluation is related to grammatical error
detection \cite{atwell1987detect,wagner2007comparative,
schmaltz2016sentence,liu2017exploiting}
and grammatical error correction \cite{islam2011correcting,
ng2013conll,ng2014conll,bryant2015far,
yuan2016grammatical}.
However, it differs from those in several aspects; most importantly,
it is concerned with the degree to which errors matter
to humans.

Work on automatic fluency evaluation in NLP has been rare.
\newcite{heilman2014predicting} predicted the fluency 
(which they called \textit{grammaticality})
of sentences written by English language learners. In contrast to ours, their approach is supervised.
\newcite{stent2005evaluating}
and \newcite{cahill2009correlating} found only low correlation between automatic metrics and fluency ratings for 
system-generated English paraphrases and the output of a German surface realiser, respectively.
Explicit fluency evaluation of NLG, including compression and the related task of summarization, has mostly been performed manually.
\newcite{vadlapudi-katragadda:2010:SRW}
used LMs for the evaluation of summarization fluency,
but their models were based on part-of-speech tags, which we do not require, and they were non-neural. Further, 
they evaluated longer texts, not single sentences like we do.
\newcite{toutanova2016dataset}
compared 80 word-overlap metrics for evaluating the content and fluency of compressions, finding only low
correlation with the latter.
However, they did not propose an alternative evaluation.
We aim at closing this gap.

\subsection{Compression Evaluation}
Automatic compression evaluation has mostly had a strong focus on content.
Hence, word-overlap metrics like ROUGE \cite{lin-och:2004:ACL} have been widely used for compression evaluation.
However, they have certain shortcomings, e.g., they correlate best for extractive compression, while we, in contrast,
are interested in an approach which generalizes to abstractive systems.
Alternatives include success rate \cite{jing2000sentence},
simple accuracy \cite{bangalore2000evaluation}, which is based on the edit distance between the
generation and the reference, 
or word accuracy \cite{hori2004speech}, the equivalent for multiple references.

\subsection{Criticism of Common Metrics for NLG}
In the sense that we promote an explicit
evaluation of fluency, our work is in line with previous criticism
of evaluating NLG tasks
with a single score produced by word-overlap metrics. 

The need for better evaluation for machine translation (MT) was expressed, e.g., by \newcite{callison2006re}, who doubted the meaningfulness of BLEU,
and claimed that a higher BLEU score was neither a necessary precondition nor a proof of
improved translation quality.
Similarly, \newcite{song2013bleu} discussed BLEU  being unreliable at the sentence or sub-sentence level
(in contrast to the system-level), or for only one single reference.
This was supported by \newcite{isabelle-cherry-foster:2017:EMNLP2017},
who proposed a so-called challenge set approach as an alternative.
\newcite{graham-EtAl:2016:COLING} performed a large-scale evaluation of human-targeted metrics for machine translation, which can be seen as a compromise between human evaluation and fully automatic metrics. They also found fully automatic metrics to correlate only weakly or moderately with human judgments.
\newcite{bojar2016ten} further confirmed that automatic MT evaluation methods do not perform well with a
single reference. 
The need of better metrics for MT has been addressed since 2008 in the WMT metrics shared
task \cite{bojar-EtAl:2016:WMT2,W17-4755}.

For unsupervised dialogue generation, \newcite{liu-EtAl:2016:EMNLP20163} obtained close to no correlation with human judgements for BLEU, ROUGE and METEOR.
They contributed this in a large part to the unrestrictedness of dialogue answers, which makes it hard to match given references.
They emphasized that the community should move away from these metrics for dialogue generation tasks,
and develop metrics that correlate more strongly with human judgments.
\newcite{elliott-keller:2014:P14-2} reported the same for BLEU and image caption generation.
\newcite{duvsek2017referenceless} suggested an RNN 
to evaluate NLG at the utterance level, given 
only the input meaning representation. 

\input{long.future_work}

\section{Conclusion}
We empirically confirmed the effectiveness of
SLOR, 
a LM score which accounts for the effects of sentence length
and individual unigram probabilities, 
as a metric for fluency evaluation of the NLG task of automatic compression at the sentence level. 
We further introduced WPSLOR, an adaptation of SLOR to WordPieces,
which reduced both model size and training time at a similar evaluation performance.
Our experiments showed that our proposed referenceless metrics correlate significantly better with fluency ratings for the outputs of compression systems
than traditional word-overlap metrics on a benchmark dataset.
Additionally, they 
can be applied even in settings where no references are available, or would be costly to obtain.
Finally, for given references, we proposed the
reference-based metric ROUGE-LM, which consists of a combination of WPSLOR and ROUGE.
Thus, we were able to obtain
an even more accurate fluency evaluation.

\section*{Acknowledgments}
We would like to thank Sebastian Ebert and Samuel Bowman for their detailed and helpful feedback.

\bibliography{conll2018}
\bibliographystyle{acl_natbib_nourl}

\end{document}

%% file: long.future_work.tex
\section{Future Work}
The work presented in this paper brings up multiple
interesting next steps for future research.

First, in Subsection \ref{subsec:ana_domain}, we investigated the performances of WordSLOR and WPSLOR in dependence of the domain of the considered text. We concluded that an application was possible even for unrelated domains. However, we did not experiment with alternative LMs, which leaves the following questions unresolved: (i) Would training LMs on specific domains improve WordSLOR's and WPSLOR's correlation with human fluency judgments, i.e., to what degree are the properties of the training data important? (ii) How does the size of the training corpus influence performance? 
Ultimatly, this research
could lead to answering the following question: Is it better to train a LM on a small, in-domain corpus or
on a large corpus from another domain?

Second, we showed that, in certain settings, Pearson correlation does not 
give reliable insight into a metric's performance. Since in general evaluation of \emph{evaluation metrics} is hard, an important topic for future research 
would be the investigation of better ways to do so, or to study under which conditions a metric's performance can be assessed best.

Last but not least, 
a straight-forward continuation of our research would encompass the investigation of SLOR as a fluency metric for other NLG tasks or languages. While the results for compression strongly suggest a general applicability to a wider range of NLP tasks, this has yet to be confirmed empirically. As far as other languages are concerned, the question what influence a given language's grammar has would be an interesting research topic.